\documentclass{article}

\usepackage{arxiv}

\usepackage[utf8]{inputenc} 
\usepackage[T1]{fontenc}    
\usepackage{hyperref}       
\usepackage{url}            
\usepackage{booktabs}       
\usepackage{amsfonts}       
\usepackage{nicefrac}       
\usepackage{microtype}      
\usepackage{lipsum}		

\usepackage[font=footnotesize]{caption}
\usepackage{amsmath}
\usepackage{graphicx}

\title{Coloring the Black Box: visualizing neural network behavior with a self-introspective model}

\author{
 Arturo Pardo \\
Photonics Engineering Group\\
Universidad de Cantabria\\
and Instituto de Investigación Valdecilla (IDIVAL)\\
39005 Santander, Cantabria, Spain\\
\texttt{arturo.pardo@unican.es} \\
   \And
 Jos\'e A. Guti\'errez-Guti\'errez\\
Photonics Engineering Group\\
Universidad de Cantabria\\
and Instituto de Investigación Valdecilla (IDIVAL)\\
39005 Santander, Cantabria, Spain\\
\texttt{gutierrja@unican.es} \\
\AND
Jos\'e Miguel L\'opez-Higuera \\
Photonics Engineering Group,\\
(Universidad de Cantabria)\\
IDIVAL, and CIBER-BBN \\
39005 Santander, Cantabria, Spain\\
\texttt{lopezhjm@unican.es} \\
   \And
Brian W. Pogue\\
Thayer School of Engineering\\
Dartmouth College\\
Hanover, New Hampshire, 03755\\
\texttt{brian.w.pogue@dartmouth.edu} \\
   \And
Olga M. Conde\\
Photonics Engineering Group,\\
(Universidad de Cantabria)\\
IDIVAL, and CIBER-BBN \\
39005 Santander, Cantabria, Spain\\
\texttt{olga.conde@unican.es} \\
}

\begin{document}
\maketitle

\begin{abstract}
The following work presents how autoencoding all the possible hidden activations of a network for a given problem can provide insight about its structure, behavior, and vulnerabilities. The method, termed \textit{self-introspection}, can show that a trained model showcases similar activation patterns (albeit randomly distributed due to initialization) when shown data belonging to the same category, and classification errors occur in fringe areas where the activations are not as clearly defined, suggesting some form of random, slowly varying, implicit encoding occurring within deep networks, that can be observed with this representation. Additionally, obtaining a low-dimensional representation of all the activations allows for (1) real-time model evaluation in the context of a multiclass classification problem, (2) the rearrangement of all hidden layers by their relevance in obtaining a specific output, and (3) the obtainment of a framework where studying possible counter-measures to noise and adversarial attacks is possible. Self-introspection can show how damaged input data can modify the hidden activations, producing an erroneous response. A few illustrative are implemented for feedforward and convolutional models and the MNIST and CIFAR-10 datasets, showcasing its capabilities as a model evaluation framework. 
\end{abstract}


\section{Introduction}
\label{sec:introduction}
the behavior of neural networks is still an open problem in the field of deep learning. Solving the \textit{black box problem}, i.e. explaining which elements of a network are used to solve each part of a given classification problem, demands the design and implementation of new tools and theoretical frameworks that enable us to untangle the result of applying automatic differentiation in a randomly-initialized distributed model. Current methods to directly approach the \textit{black-box problem} are varied. Notable examples include the \textit{distillation} of a given network's behavior into soft decision trees \cite{Frosst2017}, and the use of a parallel neural network to synthesize \textit{text-based explanations} of the specific responses of a model \cite{Hendricks2016}. While these methods show promising results in terms of providing better descriptions of neural behavior, distillation (i.e. the transformation of a deep network into a decision tree) incurs in lower accuracies, and text-based explainers provide very little insight regarding the actual networks that are involved in training and classification. Other methods include the generation of \textit{activation atlases }via dimensionality reduction techniques such as T-distributed Stochastic Neigbor Embedding (t-SNE) and Uniform Manifold Approximation and Projection (UMAP) \cite{Maaten2008, carter2019,Zahavy2016GrayingTB}, and then using \textit{Feature Visualization} to approximate via optimization the type of input that excites each of the units in the network \cite{Olah2017, Olah2018}. Previous research studied the smooth behavior of the output layer during classification, by projecting the $n$-dimensional layer onto 2D polygon maps, with predefined deterministic functions, yet ignoring the hidden layers' states \cite{Duch2003}. In more recent work, PCA and the Mapper algorithm have been applied to the weights of a feedforward network to study the topology of the weights in hidden layers during learning \cite{Gabella2019}, showing an inherent structure of divergence and specialization that should be explained in the coming years. These methodologies, unfortunately, need to be constantly rerun for incoming datapoints, and can become rather computationally expensive. Of notable mention as well are multiple methods for salience map production in images \cite{Simonyan2013, Zeiler2015} which is a means of finding neurons that are responsible of firing when presented with very specific features, such as faces. Last but not least, \textit{network dissection} associates the relevance of units in hidden convolutional layers in the detection of specific image semantic information, such as texture, material, and color detection, with the aid of a labeled dataset \cite{Bau2018}.


All these methods succeed in either (a) simplifying the structural complexity of the trained model (model induction), (b) generating human-readable sets of input images that activate specific regions of the network (deep explanations), or (c) discovering the features in the input that produce specific activations at specific hidden layers. While these are fundamental breakthroughs in the field of explainable artificial intelligence (XAI), an ideal future objective is to not only produce atlases of possible input representations that activate specific layers, but instead understanding how different activation sequences result in specific classification results. Model evaluation is, then, of paramount importance if we wish to understand when these models can be trusted, why they fail, and how they can be corrected \cite{darpa2016}. 

This article attempts to generate a geometric visualization of the domain of all possible activations in a given network with the objective of assessing its behavior. We will refer to this domain as the \textit{activation space} of a neural network. We propose the use of autoencoders (in this case, Variational Autoencoders or VAEs) to encode the hidden layers of a given network into a low-dimensional representation \cite{Rezende2014, Kingma2013}. Additionally, we show how using a neural network to generate a representational space allows us to visualize and estimate the expected errors for a validation set, provide confidence intervals, and observe its behavior against noise and simple adversarial attacks. Third, and final, we show that using an autoencoder to generate an activation map allows us to completely dissect a neural network and disentangle units with respect to their influence on the output layer. These features represent a form of self-reflection of the model by only looking at its hidden structure and repeating patterns, hence the name of the method: \textit{self-introspection}.

This work mainly focuses on experimental results shown in the case of feedforward layers and dense layers of convolutional networks. A similar result for recurrent, convolutional and attention-based models could be potentially found, as long as there is a method to provide the internal states of interest to the unsupervised model. The rest, as can be shown in the following pages, requires no significant additional preparations.

\section{Materials and methods}
\label{sec:eminem}

\subsection{Definitions for a learning problem}

Let $X$ and $Y$ be topological spaces, namely vector spaces with natural Euclidean topology. We will consider classification problems where there exists a function $f: X\rightarrow Y$ that relates an input domain $X\subseteq \mathbb{R}^m$ with an output domain $Y\subseteq \mathbb{R}^n$. Let $T_x\subset X$ and $T_y\subset Y$ represent the input and output training sets, respectively. Let the \textit{training set} be defined by pairs of input-output pairs
\begin{equation}
T = \{ (x_i, y_i) : x_i \in T_x, y_i \in T_y\},
\end{equation}
with $|T| = N_T$ the number of available training pairs. Similarly, let $V_x\subset X$ and $V_y \subset Y$ be the input and output validation sets, respectively. Let the \textit{validation set} be 
\begin{equation}
V =\{ (x_i, y_i) : x_i \in V_x, y_i \in V_y\}.
\end{equation}
Here, $|V|=N_V$ will be the number of validation set pairs. In typical learning problems, the training and validation sets will be point clouds (compact sets) of different sizes ($N_T \gg N_V$), with no overlapping, and presenting the same degree of variability among classes. In more formal terms, consider the following properties:
\begin{enumerate}
	\item Training and validation sets do not share any data: $$T_x \cap V_x = \emptyset, \hspace{1em} T_y \cap V_y = \emptyset.$$
	\item Both input and output validation sets are similarly representative of the complete set. Since we are in Euclidean space, a way to write this would be $$d_H(T_x, X) \approx d_H(V_x, X), \hspace{1em} d_H(T_y, Y)\approx d_H(V_y, Y),$$ with $d_H$ denoting the Hausdorff distance between any two point clouds. 
	\item The training and validation sets better represent the input and output spaces $X, Y$ as the number of pairs increases: $$\lim_{N_T \rightarrow \infty} d_H(T_x, X) = \lim_{N_T \rightarrow \infty} d_H(T_y, Y) = 0,$$ $$\lim_{N_V \rightarrow \infty} d_H(V_x, X) = \lim_{N_V \rightarrow \infty} d_H(V_y, Y) = 0.$$
\end{enumerate}

Let $\hat{f}: X,\theta \rightarrow \hat{Y}$ be a model or approximation of our desired (unknown) function $f$, where $\theta$ is a set of variable parameters, and $\hat{Y} = \hat{f}[X]$ is the output domain of the model (image of $\hat{f}$). Training can be defined as solving the following optimization problem:
\begin{align}
\underset{\theta}{\text{minimize  }} & E_T(\theta) = \frac{1}{N_T}\sum_{(x_i, y_i)\in T}{\|y_i-\hat{f}(x_i)\|^2}, \label{eq:optim}\\ 
\text{subject to  }  & E_T(\theta) \ge E_V(\theta) = \frac{1}{N_V}\sum_{(x_i, y_i)\in V}{\|y_i-\hat{f}(x_i)\|^2}. \nonumber
\end{align}
where $E_T(\theta)$ denotes the \textit{training error} of the model $\hat{f}$ for a given training dataset $T$, and $E_V(\theta)$ represents the \textit{validation error} of the model. The restriction of the optimization problem represents the \textit{overfitting} of the model to the training dataset, which would hinder generalization. 

To solve the optimization problem, a deep learning model will update its parameters $\theta$ through gradient descent:
\begin{equation}
\theta_t \leftarrow \theta_{t-1} - \gamma \frac{\partial E_T(\theta_{t-1})}{\partial \theta_{t-1}},
\end{equation}
and optimization will halt when $E_V(\theta_t) \ge E_T(\theta_t)$. The derivatives $\partial E_T /\partial \theta$ are calculated via automatic differentiation (also referred to as \textit{backpropagation}). Learning rate $\gamma\in\mathbb{R}^+$ can be set to a fixed value, or changed over the course of training with adaptive rate methods, such as Adam \cite{Kingma2014AdamAM} or RMSprop \cite{Graves2013GeneratingSW}.

\subsection{Notation for a feedforward network}

We will consider a multilayer perceptron with $L$ layers as a series of operations, as follows:
\begin{subequations}
	\begin{align}
	x & \nonumber\\ 
	h_0 &= \phi\left(W_0 x + b_0\right), \label{eq:net0}\\
	h_1 &= \phi\left(W_1 h_0 + b_1\right), \\
	&\vdots\\
	h_i &= \phi\left(W_i h_{i-1} + b_i\right), \\
	&\vdots\\
	\hat{y} &= \phi\left(W_L h_{L-1} + b_L\right). \label{eq:netN}
	\end{align}
\end{subequations}
In this sequence, $\phi(.)$ represents a nonlinear, continous function with a continuous inverse, $x\in\mathbb{R}^m$ receives the name of \textit{input layer}, $h_0, \dots, h_{L-1}$ are the $L$ \textit{hidden layers}, and $\hat{y}\in\mathbb{R}^n$ is the \textit{output layer}. In this model, the parameters are the weights $\{W_k\}$ and biases $\{b_k\}$. For our particular experiments, we will use ELU \cite{Clevert2015} as the main activation function for all hidden layers. 

\begin{figure}[t]
	\centering
	\includegraphics[width=0.99\linewidth]{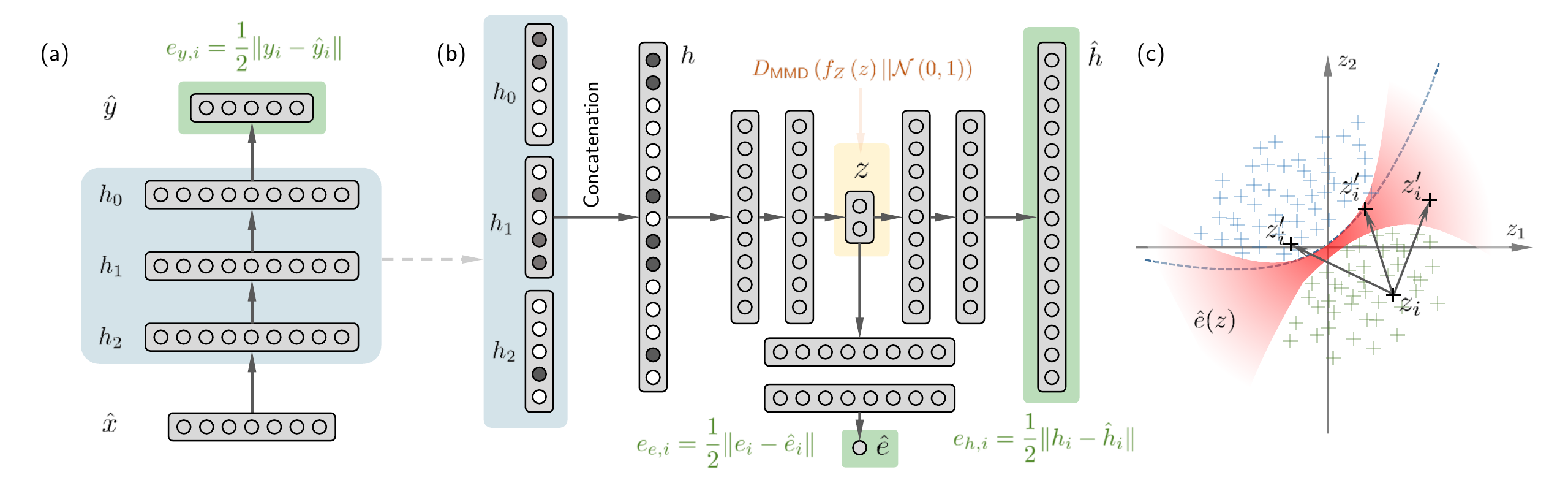}
	\vspace{-0.5em}
	\caption{Self-introspection model and main hypothesis of this work. A classifier (a) is trained on a given dataset with some general classification goal. Its hidden states represent gradual, small transformations from the input domain $x\in \mathbb{R}^m$ to the output (category) domain $y\in \mathbb{R}^n$, and thus indirectly encode the classification problem at hand. To better visualize the domain of all possible activations $h\in \mathbb{R}^p$ into a tractable embedding, an autoencoder (b) is used. Then, the autoencoder bottleneck $z$ will provide a representation of all activations in a classification problem, on top of which an estimator of the error $e = (1/2) \| y-\hat{y}\|^2$, i.e. $\hat{e}$, can be trained (subplot (c), in red). In this framework, the presence of noise, artifacts, or adversary attacks (shown as black arrows) can be identified and estimated as anomalies with respect to typical behavior for the dataset.}
	\label{fig:network}
\end{figure}

\subsection{Self-introspection}

A way to restate the \textit{black box problem} is as follows. At initialization, the weights and biases of a neural network are randomly prescribed. During training, the gradient of the cost function with respect to these weights and biases is calculated, and these parameters are updated with respect to these initial values and the activations of preceding units, which in turn also depend on random initial values and other preceding unit activations. The credit assignment of each parameter to the final classification error can be obtained via automatic differentiation due to the fact that they are randomly initialized (i.e. if they were set to zero at initialization, the gradient would be zero and no error could be back-propagated) but, as a consequence of this random initialization, the method assigns a random role to each unit. Thus, a clean relationship between the hidden activations and the output result cannot be found, even after the network has been trained.  

Informally, the activations at the hidden layers could be interpreted as representations of the input vectors as they are being converted into output vectors, as a result of applying the nonlinear equation system described by the model. This intuition can be better understood with a simpler example. If a trained feedforward classifier was only composed of one hidden layer $h$, the activations of such hidden layer directly represent the relationship between the input $x$ and the output $\hat{y}$ for any specific data pair $(x_i, y_i)$. Each specific activation pattern at the hidden layer will produce a specific response at the output layer, and will only result from a very specific set of possible input values. The hidden layer could then be sorted with respect to the amplitude of each activation for each of the output categories, and the error of the classifier could be estimated for all known hidden layer patterns.  

Under these assumptions, a hypothesis can be defined. In order to transform all the input datapoints into the same output vector, the hidden activations in a trained classifier must become more and more similar as we travel from the first to the last layers of the network. This similarity suggests that activation patterns may belong to a low-dimensional subspace or embedding, on par with the training data, and that each output unit acts as an attractor for similar, yet different data belonging to the same category. Obtaining this embedding with a secondary deep neural network yields a low-dimensional representation of all possible trajectories followed by the input data. If such trajectories belong to a low-dimensional embedding, then hidden activations should be fairly similar for input-output pairs belonging to the same category, so that the output layer yields the same result every time. Conversely, similar activation patterns will correspond to clustered points in this newly defined low-dimensional representation. 

Consider the feedforward classifier shown in Figure \ref{fig:network}.(a), with three hidden layers. If we understand the network as a distributed function approximator that gradually transforms the input vectors into output (category) vectors via Equations (\ref{eq:net0}) through (\ref{eq:netN}), minimizing Equation (\ref{eq:optim}), then the classification problem (and solution) must be encoded by the hidden activations implicitly. In order to obtain a human-readable representation of all hidden activations (at least, for the feedforward example of Figure \ref{fig:network}), the hidden layers are concatenated into a single vector, and then fed as input to an isolated autoencoder (Fig. \ref{fig:network}.(b)). The autoencoder's bottleneck will produce a low-dimensional map $z(h)$ of all possible activations, as shown in Fig. \ref{fig:network}.(c), where all the activations of the network for a given input-output pair is --in this case-- represented by a single point in 2D space. Using the bottleneck outputs as inputs to another network, an error estimator $\hat{e}(z)$ (represented by the two horizontal layers under the autoencoder's bottleneck in Fig. \ref{fig:network}.(b)) can be trained as well, by using the classification errors obtained by standard training. This is equivalent to using t-SNE or MDS, with a fundamental difference: the autoencoder serves as an explicit function that relates activations to a low-dimensional manifold and, thus, it can be trained on a given dataset, respond to incoming data and be interrogated in real time. 

This rather simplistic methodology shows interesting potential in various different problems. First, it can be used to weigh the expected accuracy of a classifier in an ensemble. Second, the representation can be used as a method for anomaly detection: uncommon activations will be placed in positions far away from common activations. Third, it allows us to sort the hidden layers of the network with respect to the expected activations for a particular output category. Finally, the autoencoder can be trained on the complete learning record of a given classifier, showing the evolution of the activation patterns as a response to data, where the network gradually specializes to solve the classification problem. 

\subsection{Notation for a self-introspecting model}

The following symbols will be used to refer to the various structures throughout the manuscript:

\begin{itemize}
	\item \textit{Classifier }$f_\text{cls}(x) = \hat{y}$. The classifier is an approximator of the function $f: X \rightarrow Y$. In particular, $f_\text{cls}: X \rightarrow H \rightarrow \hat{Y}$, where $H$ is the set of all possible hidden unit activations. It is trained by minimizing $(1/2)\| \hat{y}-y\|^2$.
	\item \textit{Autoencoder.} The autoencoder is composed of two functions: the encoder $f_\text{AE,e}(h)=z$, which provides a relationship $f_\text{AE,e}: H \rightarrow Z$; and the decoder $f_\text{AE,d}(z) = \hat{h}$, which in turns attempts to revert the encoder's operation: $f_\text{AE,d}: Z \rightarrow \hat{H}$. It is trained by the reconstruction error $(1/2)\|\hat{h}-h\|^2$ and the Maximum Mean Discrepancy (MMD) at the bottleneck.
	\item \textit{Estimator.} The estimator $f_\text{estim}(z) = \hat{e}$ generates an estimate of the error of the classifier, $e=(1/2)\cdot\|y-\hat{y}\|^2$ with the low-dimensional representation provided by the encoder. It is trained by the estimation error, namely $(1/2)\|\hat{e}(h) - e\|^2$.
\end{itemize}

These systems can be trained simultaneously or sequentially. Note that, in order to train the classifier, autoencoder and estimator independently, but at the same time, backpropagated gradients must be stopped at three locations: (a) between the hidden activation sequence and the autoencoder input, (b) between the autoencoder bottleneck and the error estimator input, and (c) between the MSE error function $e$ and the output of the network $\hat{y}$. If trained sequentially, the order must be, as can be expected, (1) classifier, (2) autoencoder, and (3) estimator.

\subsection{Model architecture}

Two classifier models were tested in the experiments. The simplest model is a feedforward $12\times200$ ELU model, which will be used for classifying MNIST samples. As an example of the same analysis on a convolutional model, an all-convolutional neural network \cite{Springenberg2014} will be tested, by evaluating the activations at the dense (feedforward) layers. The architectures of both networks are left in Table \ref{tab:classifiers}. 

In the case of the autoencoder and estimator models, the same networks were used for both classifiers. Their architectures are provided in Table \ref{tab:introspection}. The autoencoder is a feedforward InfoVAE \cite{Zhao2018}, which implements latent-space Gaussianity by the Maximum Mean Discrepancy (MMD) between the latent-space results and random samples from a Standard Normal distribution $\mathcal{N}(0,I)$. The estimator is a $3\times 200$ feedforward network with a single linear output. 

It must be noted that these architectures are mostly experimental and for illustrative purposes only. For more realistic examples, the complexity and depth of the networks shall certainly be improved. Additionally, for the convolutional case, only the feedforward hidden layers are used for self-introspection. Encoding the convolutional layers as inputs for the autoencoder should certainly be studied in further work. 

\subsection{Training routine}

The models were trained on Stochastic Minibatch Gradient Descent, with Adam as the optimization engine. Cyclic Learning Rates (CLR) \cite{Smith2015} were used with a triangular (sawtooth) learning cycle. Dropout regularization was included for all layers, with $p=0.9$, unless specified otherwise \cite{Hinton2012}. Validation errors were tested at the end of each cycle, and early stopping was implemented with a patience of $5$ cycles of monotonously increasing validation error. The complete system is implemented with Tensorflow 1.14 on Python 3.6 and run on an \textit{nVidia RTX 2080 Ti} GPU (Nvidia Corporation, Santa Clara, California, USA). 

For the feedforward network classifier and the MNIST dataset, cycle length was $T=3000$ iterations, with a learning rate range of $l_r \in [0, 10^{-5}]$. Minibatch size was 128 for the classifier, the autoencoder, and 32 for the error estimator. The total number of cycles was $N = 30$, achieving a test set accuracy of $96.57\%$. The error estimator required a total of $N=50$ cycles with a cycle length of $T=10000$. For CIFAR-10 and the all-convolutional net, cycle length was $T=5000$ and $N=100$, and dropout was set to $p=0.8$. Additionally, the maximum value in the learning rate was set to decay at a rate of $0.95$, achieving a final test set accuracy of $76.2\%$. The autoencoder is trained with an identical regime, only with $l_r \in [0, 10^{-5}]$. The error estimator is trained with twice as many cycles ($N=200$).

\subsection{Confidence metrics and error estimation}

During training, the input $x_i$ will flow through the classifier into an output category $\hat{y}_i$, producing an activation sequence $h_i$. The latter will be read by the autoencoder, which in turn will produce a representation of the activation sequence $z_i$. Then, the estimator will use $z_i$ to provide a number which should represent the expected error during training for a specific hidden layer pattern sequence, $\hat{e}(z_i)=\hat{e}(h_i)$. 

While these are still estimates at training time, as long as overall accuracy and errors are within the same order of magnitude for the training and validation steps, by avoiding overfitting as much as possible, we could consider them an adequate measure of how much a sequence activation is expected to provide an accurate response. For this particular model, the output unit at the estimator is a linear unit, which must converge to the logarithm of the error at training, $\hat{e} \approx \log_{10}(e)$. Confidence could be measured as the negative of the logarithm of the error, namely $-\log_{10}(e)$.

\setlength{\tabcolsep}{0.5em}
\renewcommand{\arraystretch}{1.4}
\begin{table}[h]
	\center
	\caption{Classifier structures}
	\label{tab:classifiers}
	\begin{tabular}{l l}
		\hline
		\textbf{Net1. Deep feedforward ($\downarrow$)} & \textbf{Net2. All-convolutional ($\downarrow$)} \\
		\hline
		$784\times 1$ Input & $32\times 32\times 3$ Input \\
		$12\times200$ ELU & $2\times 96$ ELU, f.s. $3\times 3$\\
		$10$ Sigmoidal Output  & $1\times 96$ ELU, f.s. $3\times 3$, str $=2$ \\
		& $2\times 192$ ELU, f.s. $3\times 3$\\
		& $1\times 192$ ELU, f.s. $3\times 3$, str $=2$ \\
		
		& $1\times192$, global avg.\\
		& $3\times 512$ ELU \\
		& $10$ Sigmoidal Output\\
		
		\hline
	\end{tabular}
	\begin{tabular}{p{8 cm}}
		\footnotesize{Notation: f.s. stands for \textit{filter size} (per neuron). Layer notation is $c\times n$, $c:=$ number of identical layers, $n:=$ number of hidden neurons in each layer. Finally, \textit{str} indicates striding. Flattening and reshaping layers implemented via \texttt{tf.contrib.layers.flatten()} and \texttt{tf.reshape()}.}
	\end{tabular}
\end{table}

\setlength{\tabcolsep}{0.5em}
\renewcommand{\arraystretch}{1.4}
\begin{table}[h]
	\center
	\caption{Self-introspection models}
	\label{tab:introspection}
	\begin{tabular}{l l}
		\hline
		\textbf{Feedforward InfoVAE ($\downarrow$)} & \textbf{Feedforward estimator ($\downarrow$)} \\
		\hline
		$N_h\times 1$ Input & $2$ Input \\
		$200$ ELU & $200$ ELU\\
		$200$ ELU & $200$ ELU\\
		$200$ ELU & $200$ ELU\\
		$2\times1$ Linear & 1 Linear Output \\
		$200$ ELU & \\
		$200$ ELU & \\
		$200$ ELU & \\
		$N_h\times 1$ Linear Output & \\
		\hline
	\end{tabular}
	\begin{tabular}{p{8 cm}}
		\footnotesize{Layer notation is $c\times n$, $c:=$ number of identical layers, $n:=$ number of hidden neurons in each layer. }
	\end{tabular}
\end{table}

\subsection{Reorganizing randomly initialized layers}
\label{sec:reorganize}
Let us consider a self-introspective system consisting of a feedforward classifier $\hat{y}= f_\text{cls}(x)$, an autoencoder composed of an encoder $z=f_\text{AE,e}(h)$ and a decoder $\hat{h} = f_\text{AE,d}(z)$ (where here $h$ is the concatenation of all hidden layers in the classifier) and an error estimator $\hat{e}=f_\text{estim}(z)$. We will assume that we can split a training dataset $T$ into the various categories they belong to, i.e. $T_0 , \dots, T_K\subseteq T$, with $T_i = T|H_i$ representing the subset of $T$ that belongs to the $i$-th class. Since $T$ can be understood as a set of samples from a continous random variable $\mathcal{X}$, then the vector of hidden states can also be defined as a continuous random variable $\mathcal{H}=g(\mathcal{X})$, related to the input by $g(.)$, which can be deduced from Equations (\ref{eq:net0}) through (\ref{eq:netN}). Let the domain of $\mathcal{H}$ be $\Omega$, and let its joint probability density function (PDF) for each of the elements of $h=(h_0, \dots, h_N)^T$ given hypothesis $H_k$ be $f_{\mathcal{H}}(h|H_k)=f_{\mathcal{H}}(h_0, \dots, h_N | H_k)$. The \textit{expected hidden activation pattern} of the network given a type of hypothesis or category of input data $H_k$, $\mathbb{E}[h|H_k]$, could in theory be calculated via the multiple integral
\begin{equation}
\mathbb{E}[h|H_k] = \int_{\Omega}{h \hspace{0.2em} f_{\mathcal{H}}(h_0,\dots, h_N|H_k) \hspace{0.2em} dh_0\dots dh_N}.
\end{equation}

This volume integral becomes quickly computationally expensive as the number of units increases, and becomes intractable if the domain of the random variable is not bounded, as is the case for ELU units, or if the input data is highly dimensional, making it difficult to obtain $f_\mathcal{H}(h|H_k)$ if the PDFs need to be estimated.

If an autoencoder is used to obtain $\mathcal{Z} = f_\text{AE,e}(\mathcal{H})$, then this random variable has a tractable PDF or \textit{prior} approximately equal to the unit Gaussian $f_\mathcal{Z}(z) =\mathcal{N}(0, I)$. Similarly, the expected bottleneck values, or expected value of random variable $\mathcal{Z} =(Z_1, Z_2)$ for a specific hypothesis $H_k$ can be calculated much more quickly via 
\begin{equation}
\mathbb{E}[z|H_k] = \int_{\Omega_z}{z \hspace{0.2em} f_{\mathcal{Z}}(z_1, z_2|H_k) \hspace{0.2em} dz_1 dz_2},
\end{equation}
where here $\Omega_z$ is still an unbounded domain, but it can be truncated to a smaller domain $\Omega'$ (e.g. $\Omega' = [-4, 4]\times [-4, 4]$) with negligible accuracy losses, since $f_Z(z)$ is only nonzero near the origin. With this in mind, the previous integral can be numerically approximated with a Riemannian sum:
\begin{equation}
\mathbb{E}[z|H_k] \approx \sum_{i=-N}^{N}{\sum_{j=-M}^{M}{z_{ij} \hspace{0.3em} \hat{f}(z_{ij}|H_k) \hspace{0.3em} (\Delta z)^2}},
\end{equation}
with $z_{ij} = (\Delta z \cdot i, \Delta z \cdot j)^T, \hspace{0.2em} i=-N, \dots, N, \hspace{0.2em} j=-M, \dots, M$ and $\hat{f}(z|H_k)$ is an estimate of the joint PDF of $\mathcal{Z}$, which can be obtained with the training data. With this estimate, the decoder can be used to obtain an estimate of the activation pattern represented by that low-dimensional point:
\begin{equation}
\mathbb{E}[h|H_k] \approx f_\text{AE,d}\left(\mathbb{E}[z|H_k]\right).
\end{equation}

The network's units can be sorted by their expected activation to all hypotheses, via
\begin{equation}
\hat{d_i} = \underset{k}{\arg \max} \hspace{0.3em} \mathbb{E}[h_i|H_k].
\label{eq:max_expected}
\end{equation}
In this way, each unit is assigned to the dataset category to which they are most sensitive. If a unit serves more than one purpose, it will at least be assigned to the most significant category. This operation results in a sorting of all layers with respect to the output neuron that they activate with.

\subsection{Artificial brainbow}

One way to visually identify how much each unit in a layer contributes to a specific output classification is by color-coding each of them in terms of the expected intensity of each activation for each type of output, $\mathbb{E}[h_i|H_k]$. This can be easily done with a series of colors $c_0, \dots, c_K$, coded as 3-dimensional RGB vectors, and associated for each of the output categories $H_0, \dots, H_K$, and then calculating:
\begin{equation}
\alpha_i = \frac{1}{K}\sum_{k=1}^{K}{c_i \cdot \mathbb{E}(h_i|H_k)}.
\label{eq:colorcoding}
\end{equation}
The units will then acquire a more grayish tone when not supporting any specific output category, or take the color of the output category if the opposite is true.

\subsection{Robustness against noise and adversarial attacks}
\label{sec:adversarial}

We will study how the autoencoder interprets the activations of the network with a series of simple experiments. First, noise will be added to MNIST on a model that has been trained on clean samples. The influence of noise on the internal states of the network should be percievable enough to produce a visible representation. Then, the network can be trained under noise, and its performance can be re-evaluated. Since the latter network has been prepared to operate under noise, its responses should be more consistent than the network that was trained with clean samples, which should then be reflected by their accuracy.  

Similarly, the network's resilience to adversarial attacks can be studied as trajectories in activation space. An adversarial attack consists essentially in a small variation (also known as a perturbation) in the input data that produces significantly different activation sequences in a classifier, resulting in an erroneous result by a large margin. For these experiments, the Fast Gradient Sign Method (FGSM) will be used to modify the internal states of the network from the input \cite{Goodfellow2014}. A successful attack shall then be defined as a trajectory in activation space from the original (correct) category to the target (wrong) category, whereas the opposite will be true for failed attacks.

\section{Results and discussion}
\label{sec:results}

The results are divided in 6 sections, dedicated to showcasing some practical results of using an autoencoder on the hidden activations. Our main objective is to present the flexibility and interactivity of these autoencoders in state-of-the-art problems.

\subsection{Examples of activation spaces}

Figure 2 shows the visualizations that can be achieved by autoencoding hidden activations. The top row corresponds to \textit{Net1}, the feedforward classifier trained for MNIST, and the bottom row shows the dense activations of \textit{Net2}, its convolutional counterpart, trained for CIFAR-10. These plots represent the overall behavior of the network when provided with various different inputs, empirically showing that (a) there exists a low-dimensional embedding where hidden activations in a deep model are similar for similar inputs, that (b) that a 2D representation of this embedding where sufficient separability across activation patterns can be achieved; and that (c) wrongly classified inputs are a product of uncommon hidden activations.

\begin{figure*}[t]
	\centering
	\includegraphics[width=1.0\linewidth]{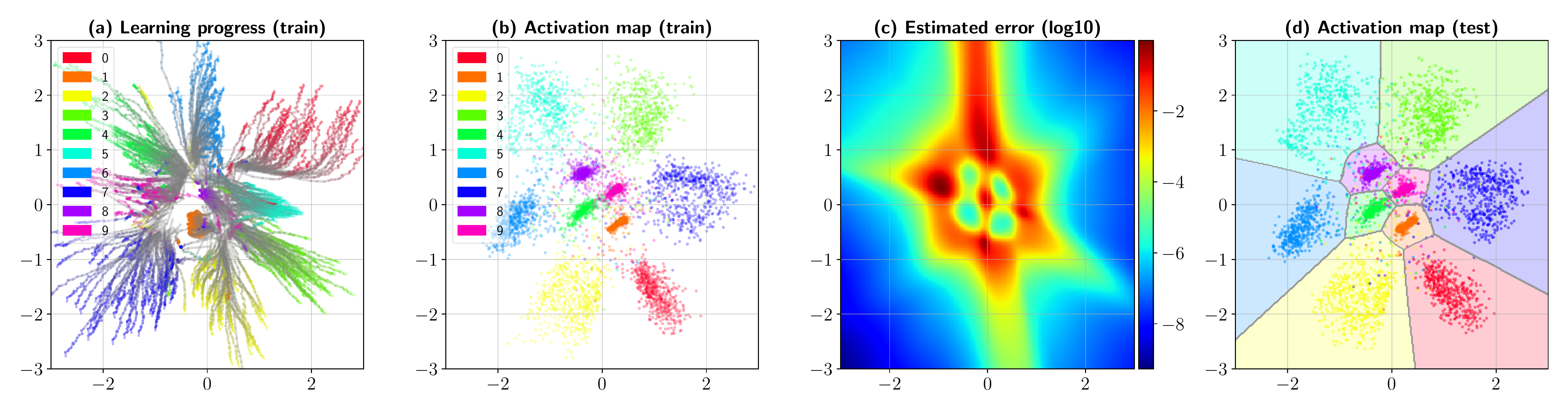}
	\includegraphics[width=1.0\linewidth]{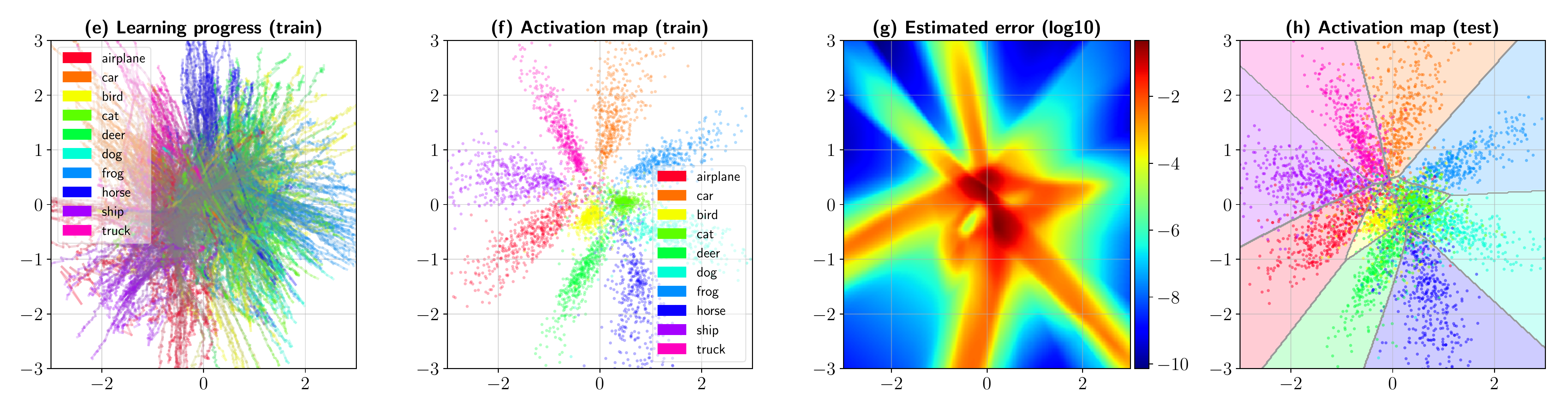}
	\vspace{-1.5em}
	\caption{Training and classification errors seen through self-introspection. Top row: (a) Activation patterns shown by hidden layers of \textit{Net1} on MNIST during training can be stored and then used to train an autoencoder, showing the divergence over time in hidden layer activations as the units specialize. Similarly, a different autoencoder can be trained on the activations on the already trained classifier (b). Once convergence is achieved, the estimated error during training can be learned (c), and the activations for the test set can be observed (d). Bottom row, (e) through (h): ditto, for \textit{Net2} and dataset CIFAR-10. Misclassification occurs when the input produces an activation pattern in between well-known activations with clear responses. Note: in these plots the $x$ and $y$ axes correspond to units $z_1$ and $z_2$ at the VAE bottleneck, respectively, and have been omitted to save space.}
	\label{fig:act_map}
\end{figure*}

As a first experiment, the hidden states of the network were recorded over a total of 30 cycles, or CLR epochs. Then, the complete record was used to train a VAE, so that the trajectories of the hidden activations with respect to the inputs could be plotted. The result of this calculation is shown in Figure \ref{fig:act_map}.(a), where each line represents a handwritten digit (0--9). The trajectories fade from grey into the color of the ground truth category they belong to, in order to better depict the passage of training time. In this representation, it is possible to observe how successive learning cycles separate further the hidden activations that correspond to each of the categories. Once convergence is achieved, a new VAE can be used to encode the final activation patterns for the complete MNIST dataset (Figure \ref{fig:act_map}.(b)). In this subfigure, each of the input datapoints is color-coded by ground truth as well. Misclassified digits can be seen as input datapoints that elicit an activation pattern that is sufficiently abnormal to result in an erroneous output response. By studying the mean square error (MSE) of the output layer during training (Figure \ref{fig:act_map}.(c)), we can observe how the misclassified digits correspond to activation patterns in-between clearly defined responses, for both the training set and the test set, the latter shown in Figure \ref{fig:act_map}.(d). A similar result is achieved by autoencoding the densely connected layers of \textit{Net2}, in Figs. \ref{fig:act_map}.(e) through (h). 

Figures \ref{fig:act_map}.(a) and \ref{fig:act_map}.(e) showcase how some of the datapoints become more separated than others during training. This can be thought of as a result of the training process, where some inputs cannot be separated as easily as other, more clearly-defined ones. Hidden activations in the classifier are varied for a given category, yet they all seem to cluster together in a family of similar activation patterns (or data trajectories) that produce the same output. Datapoints that elicit unclear activations lie in the fringes of well-defined clusters, thus probabilistically resulting in misclassification after passing the output sigmoidal/softmax layer through an $\arg \max$ operator to obtain the output category.

The classification error estimates in Figures \ref{fig:act_map}.(c) and \ref{fig:act_map}.(g) depict clear spatial corregistration between the estimated mean square error (MSE) and the locations of the activation patterns in $z$-space (i.e. the space defined by the bottleneck of the autoencoder) that produce classification errors, for both the training and test sets. By using this estimate of the error in a confidence metric $c=-\log_{10}(\hat{e})$, unclear classifications could be discarded or detected as anomalies, which could be useful in classification tasks where reliability metrics are a necessity.

\subsection{Reordering a feedforward network}

Being able to calculate the expected activation patterns of a given network and obtaining a low-dimensional embedding of its behavior not only provides a functional visual representation, but may also serve as a tool for dissecting and reordering the layers of a network, as well as studying the paths that the input data take as they merge and become the output vector. 

\begin{figure*}[t]
	\centering
	\includegraphics[width=1.0\linewidth]{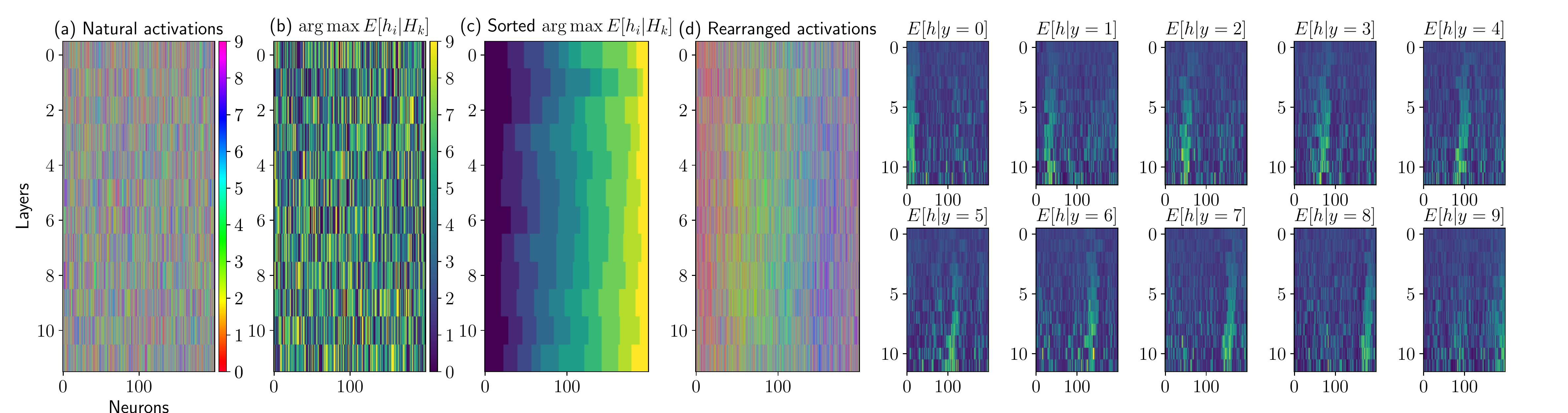}
	\vspace{-1.5em}
	\caption{Summary of a neural network rearrangement process. (a) The original arrangement of the units in each layer is random due to initialization; here, the expected activation $\mathbb{E}[h|H_k]$ associated with each neuron is color-coded as per Equation \ref{eq:colorcoding}. The $H_k$ that elicits the strongest $\mathbb{E}[h_i|H_k]$ in each unit is then obtained (b), which allows for the layer-wise sorting of each of the hidden layers (c). While it is possible to color-code them again to study their structure (d), another interesting feature is observing the expected activation patterns of the hidden units for each of the digits in MNIST (10 blue plots on the left). In these plots, each of the layers is normalized to 1. Each of the input datapoints seem to, on average, follow a funnel-like structure inside the unraveled network, which slowly turns the vector into the output category.}
	\label{fig:rearranged}
\end{figure*}

The following result is presented for the feedforward network \textit{Net1} only, mostly for illustrative purposes. Figure \ref{fig:rearranged} depicts the complete process, as described by the equations in Section \ref{sec:reorganize}. Figure \ref{fig:rearranged}.(a) provides a representation of the affinity of each unit to their respective output category, via color-coding (Equation \ref{eq:colorcoding}). In these plots, Layer 0 corresponds to the hidden layer closest to the input, and Layer 11 directly precedes the output layer. As per Equation \ref{eq:max_expected}, trained units seem to have a preferred output category, i.e. a set of similar inputs from the same category that elicit a maximal strength of activation. This allows us to sort each of the layers in terms of the type of output category that they are most likely to respond to (Fig. \ref{fig:rearranged}.(c)). This results in a reordered network (Fig. \ref{fig:rearranged}.(d)), where data takes different paths depending on their actual, ground truth category (ten right subplots in Fig. \ref{fig:rearranged}). 

A few things can be noted about the behavior of the rearranged network. First, Fig. \ref{fig:rearranged}.(c) provides insight about the relative relevance of each output category in the hidden layers: the relative distribution of units dedicated to each category is more or less uniform. This could be explained by the fact that MNIST is a balanced dataset, and thus similar amounts of the network are dedicated to activating when presented with each of the 10 different handwritten digits. Secondly, it appears as if each input digit follows a very specific path inside the network, proving (at least empirically) that there exists a common activation pattern (albeit with slight variations to accept input variability) for every output category. These \textit{activation columns}, seen as bright yellow streaks that change for each category, show some degree of overlapping, which could potentially be due to dropout regularization. Additionally, while many of the units are randomly activated, an exact subset of them (as detected by the sorting method in (c)) activate more than others, while another fraction (less noticeable but color coded in dark blue near the bright columns in (e) through (o)) of the network is dedicated to being inhibited. This seems to suggest that, within a deep neural network, there exist a defined subset of units dedicated to excitation, another fraction dedicated to inhibition, and a much larger section that is orthogonal (i.e. showcasing near zero activation) to each of the classification categories. A much more scattered, weaker fourth subset is subtly activated across all categories, likely corresponding to units with more than one specific role in the classification problem. These fractions and elements are statistically the same for a given classification category, and can now be observed and sorted, suggesting there is a clear latent structure within a deep classifier, which unfortunately could not be observed before due to random initialization.

\subsection{Ordered activation sequences and latent representations}

\begin{figure*}[t]
	\centering
	\includegraphics[width=1.0\linewidth]{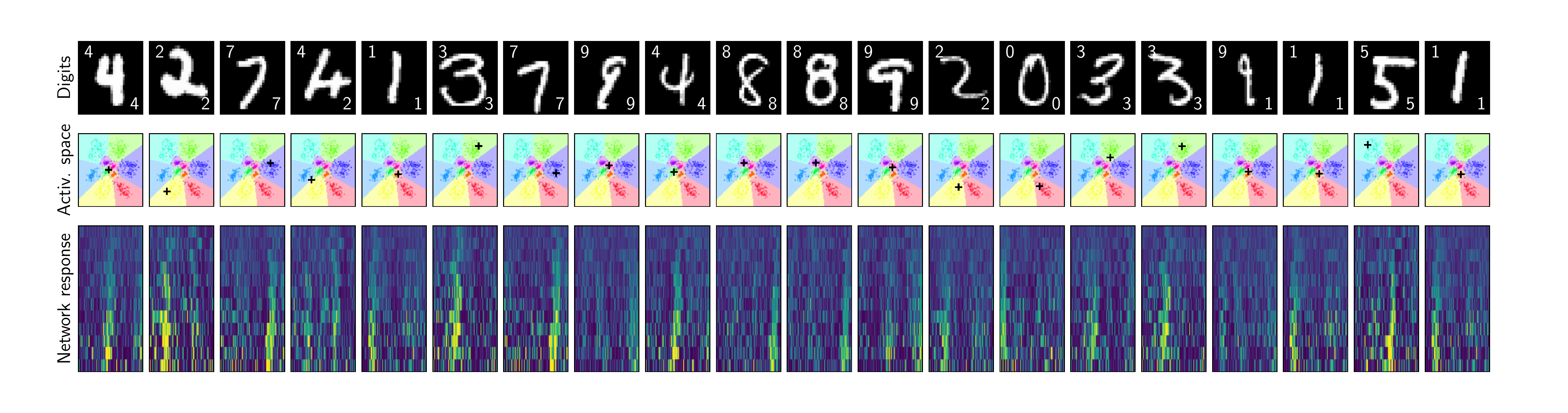}
	\includegraphics[width=1.0\linewidth]{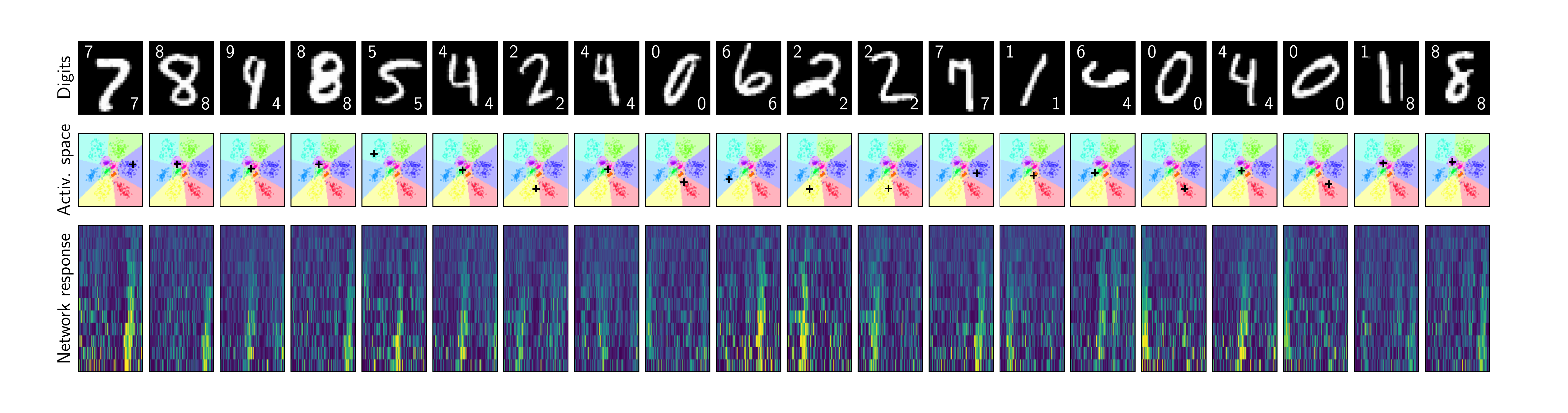}
	\vspace{-2.0em}
	\caption{Two rows of visualization examples for MNIST and the untangled feedforward network. Each handwritten digit (top dark images) also show the actual digit value (number at the top left corner) and the classifier's decision (bottom right corner). Their corresponding activation pattern in the ordered network is shown in both activation space (center subplots) and visualized directly (bottom subplots). Misclassified digits correspond to less intense activations or multiple simultaneous sub-graphs being activated at the same time, which corregister with locations in activation space that showcase high average classification errors (the current activation pattern of the network for the given digit is plotted as a black cross in activation space). The color range for all  network response images is fixed to $[-1, 4]$.}
	\label{fig:examples}
\end{figure*}

To study the correspondence of pattern activations with the classification input and the latent space, a few examples of activation patterns for a series of MNIST digits are shown in Figure \ref{fig:examples}. Two sets of 20 samples are displayed. Each number can be represented in three different ways: a black-box perspective (top plots for each digit), where the handwritten digit (center), its actual category (top left number) and the classifier's output (bottom right digit) are provided, allowing us to directly identify misclassification; an activation-space representation of the current state of the network, which is presented as a black cross in the activation map; and direct observation of the rearranged classifier activations. The two latter allow us to study the output activation sequence further. In general terms, there seems to be a substantially clear relationship between the strength and clarity of the activation pattern and its location in activation space. Stronger (i.e. more intense) activation patterns tend to be embedded in clusters that are far away from fringe areas, whereas weaker, noisier activations that correspond to misclassifications will stray away from such clusters, and exist between two or more of them, resulting in multiple parts of the network being activated in such a way that the hard classification output becomes corrupted. Let us comment two examples: the fourth digit from the left in the first row, the 4 classified as a 2, elicits a weak hidden activation sequence that is interpreted by the autoencoder as intermediate between the $y=4$ and $y=2$ output categories. Similarly, the sixth digit from the right in the second row, a 6 misclassified as a 4, elicits two responses in the network: it activates regions associated mostly with $y=6$ and with $y=4$ at the same time. The strongest one, possibly due to its location and small loop size, turns out to be $y=4$. In activation space, this is translated as a hidden activation sequence that belongs to neither typical activations for $y=6$ and $y=4$, and so the sequence rests halfway between the typical clusters associated with those output categories. 

We can empirically conclude that, after taking appropriate measures to avoid overfitting, and assuming sufficient variety and representativity of the true distributions in the training dataset, abnormal behaviors will exist in between clusters representing defined, strong activation patterns and, thus, may be detectable and corrected, insofar as the corruption of the input information is not severe to the point of producing an activation sequence typical of another category. More research is needed to study the presence of outliers and extreme adversarial examples, as well as providing a theoretical model that can explain these results. The following subsections will attempt to provide illustrative examples of the applicability of this method for some current problems in deep learning.

\subsection{Dataset noise and network performance}

\begin{figure*}[t]
	\centering
	\includegraphics[width=1.0\linewidth]{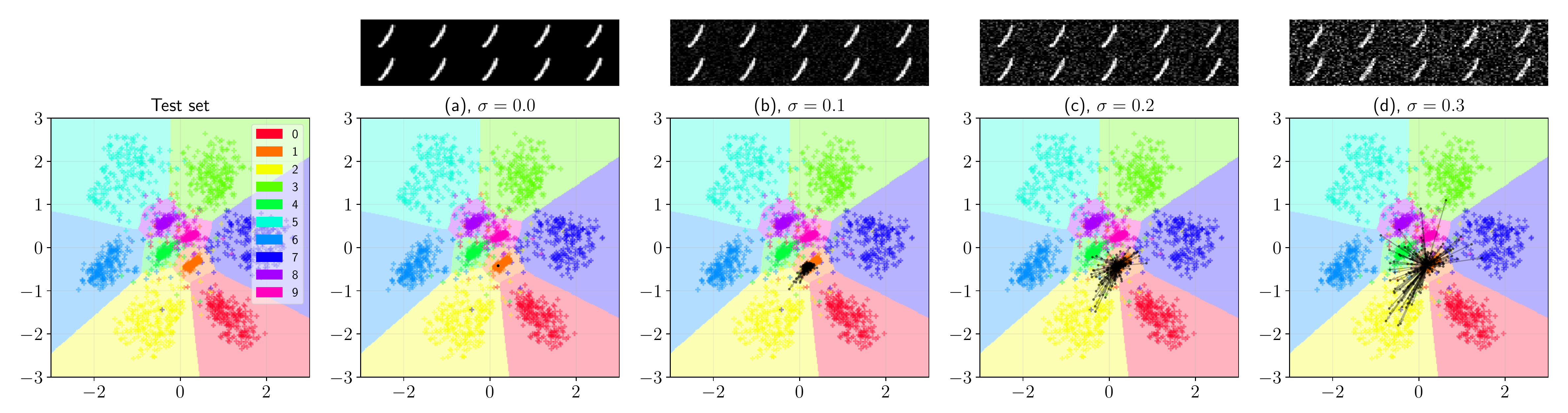}
	\includegraphics[width=1.0\linewidth]{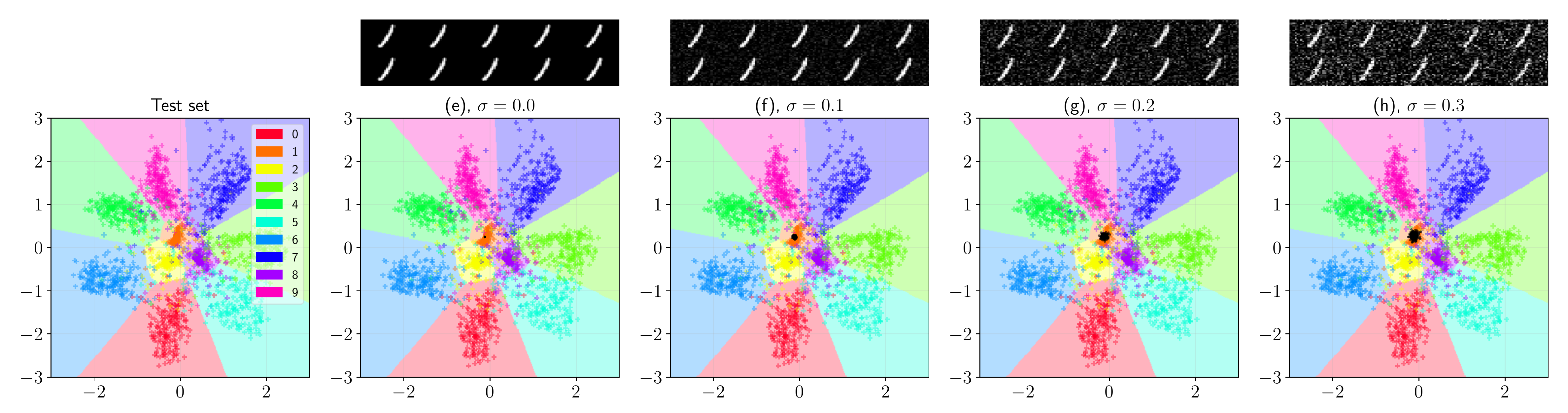}
	\vspace{-2.0em}
	\caption{Resilience to input noise improvements via noise injection can be visualized via self-introspection. Constellation diagrams (a) through (d) depict the behavior of \textit{Net1} to Sample \#907 in the MNIST test set with various levels of input noise power, from $\sigma=0$ to $\sigma=0.3$. Samples of noisy inputs are provided above each of the subplots. The network is fed with 200 inputs of Sample \#907 with additive white Gaussian noise (AWGN). Black arrows in the 2D activation map represent the displacement of the hidden activation pattern from the original response. In (e) through (f), the activation map of an identical network trained with noise injection and domain randomization ($\sigma \in [0, 1]$) of the AWGN power is shown, portraying a much more stable response to noise, with less significant classification errors. Each network's original responses to the noiseless MNIST test set are provided at the leftmost subplot for reference.}
	\label{fig:noise}
\end{figure*}

A first problem, which is easy to study, consists in visualizing the performance of a network when trained with clean data and then evaluated with data that contains noise. Such a representation would be analogous to a constellation diagram of PSK and QAM signals in digital communication systems, where symbol detection errors in a channel can be visualized as displacements from the reference input signals in a 2D map \cite{Etten2005appA}. While the overall accuracy as a function of input noise variance $\sigma^2$ can be studied as a black box, the representations provide an additional picture where we can observe the damage caused to the hidden layers' response as noise power increases. The experiment is summarized in Figure \ref{fig:noise}. First, a feedforward network was trained on the MNIST training set. Then, a single test set digit is selected (for Fig. \ref{fig:noise}, Sample 907 was chosen, a handwritten number one). To generate the noisy input test set, 200 samples of additive white Gaussian noise (AWGN) with size $28\times 28$ are generated and added to identical copies of Sample 907. This set is fed into the classifier, and its behavior is recorded. The recorded hidden states are then given to the autoencoder, who generates 200 representations in the 2D activation map. Very much like in quadrature amplitude modulation (QAM) constellation diagrams, the network's response strays away from its original response as input noise power is increased. This deviation is represented in the subplots (a) through (d) for increasing values of input power for \textit{Net1}, who has never experienced AWGN noise before, and shows a preference to misdetect noisy ones as twos and fours for very small noise variance (note \ref{fig:noise}.(c), for instance). This preference could be thought of as a susceptibility for adversarial noise in a specific direction in activation space.

Secondly, a clone of \textit{Net1} is trained again for the same number of epochs, but this time noise is injected at the input layer (AWGN with $\sigma=1.0$). To improve its capability to classify under all sorts of input noise power up to a limit, domain randomization is included in the training routine, so that $\sigma$ can vary from $0$ (noiseless classification) to $\sigma=1$ (maximum noise) for each minibatch. Its activation space map (or its classification constellation diagram) is plotted in Figs. \ref{fig:noise}.(e) through \ref{fig:noise}.(h), presenting a much more stable response to Sample \#907 to various levels of input AWGN power. This observable improvement in generalization power by introducing noise in the model has been shown empirically before \cite{Zur2009, Rakin2018}, by studying the overall accuracy of the network. The presented observations suggest that noise injection may be modifying the type of implicit function that the network finds optimal for minimizing the cost function. After all, the noise-injected network responds similarly to inputs that produce radically different output responses in \textit{Net1}.

\subsection{Representation of adversarial examples}

\begin{figure*}[t]
	\centering
	\includegraphics[width=0.99\linewidth]{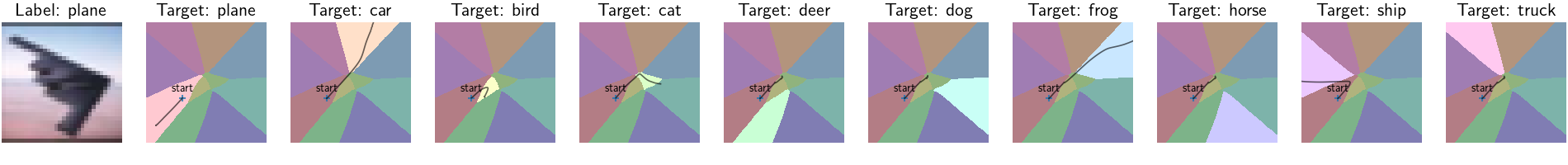}
	\includegraphics[width=0.99\linewidth]{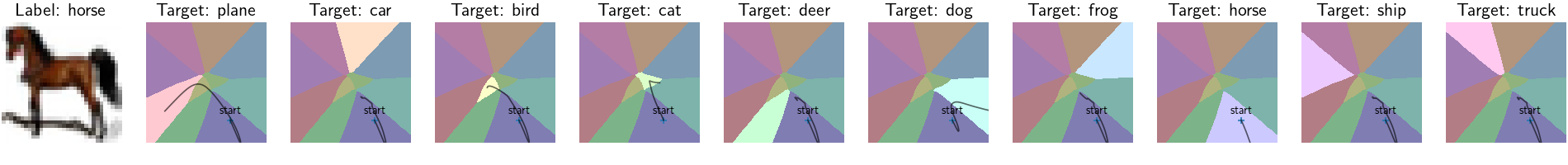}
	\includegraphics[width=0.99\linewidth]{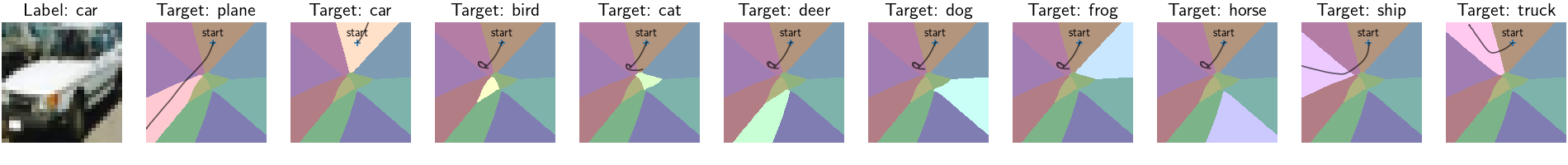}
	\includegraphics[width=0.99\linewidth]{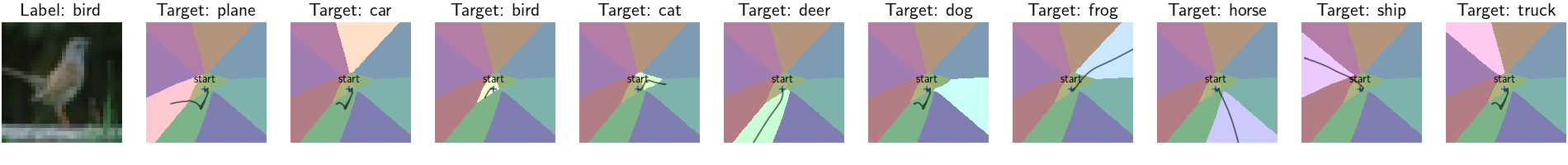}\\
	\includegraphics[width=0.99\linewidth]{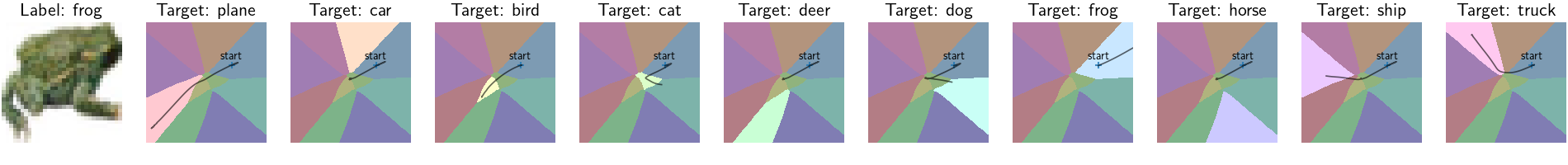}
	\includegraphics[width=0.99\linewidth]{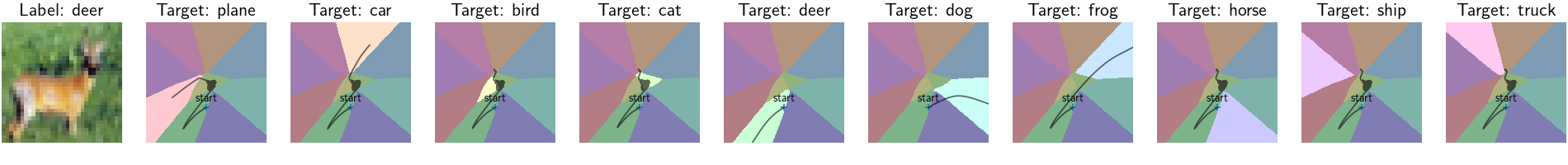}
	\includegraphics[width=0.99\linewidth]{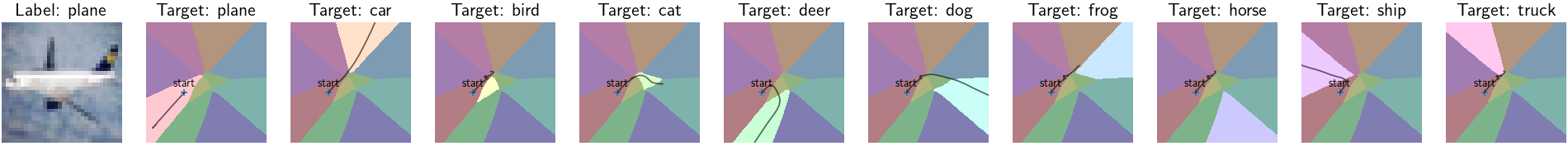}
	\caption{Visualizations of adversarial attacks with the Fast Gradient Sign Method (FGSM). These attacks correspond to CIFAR-10 test images, classified by a trained \textit{Net2}. FGSM parameters were $\epsilon=0.01$ and $N=100$ iterations, with the objective of showing the evolution of the hidden activations as the input image is modified. The initial activation pattern is marked with a cross and the text \textit{'start'}. Once the adversarial attack begins, the hidden activation patterns follow a trajectory implicitly controlled by the gradient $\partial E_\text{adv}/\partial x$. Not all adversarial attacks succeed in 100 steps, depending on the initial hidden activation pattern and input image.}
	\label{fig:adv_attacks}
\end{figure*}

Having a visual representation of all possible activations may also be helpful in the understanding of adversarial attacks. As defined in Section \ref{sec:adversarial}, FGSM was deployed as a typical model of adversarial noise. The gradient of the cost function $E_\text{adv} = (1/2)\| y_\text{target} - \hat{y}\|^2$ for a desired output category $y_\text{target}$ was calculated, and the input $x$ was modified as to minimize $E_\text{adv}$ in a series of steps. For this particular configuration of FGSM, the number of steps was 100 and $\epsilon = 0.01$. The experiments were prepared on a trained \textit{Net2}. Figure \ref{fig:adv_attacks} provides six examples from the test set in CIFAR-10. Each of the subplots in every row represents one of the ten possible adversarial attacks for this dataset. 

A few interesting conclusions can be drawn from these visualizations. First, the reader can notice that the trajectories followed in activation space are different for each input image, but there are common patterns amongst them. A successful adversarial attack occurs when the sequence of hidden activations is modified in such a way that the output layer provides the target category instead of the correct one. In activation space, this is visualized as a trajectory that begins in a point corresponding to an adequate activation sequence and then gradually becomes a hidden activation sequence that provides the desired, wrong classification. In many cases, these paths are traveled by more than one adversarial example, suggesting that certain activation sequences represent some sort of gateway to very specific activations. Second, not all FGSM attacks succeed after 100 steps, suggesting that some activation steps and inputs are more resilient to adversarial attacks than others. For example, in the third from top case of Figure \ref{fig:adv_attacks}, the network can be easily convinced that the white car is in fact a ship, a plane, or a truck, but is resilient to wrongly identifying the image as a bird, a deer, a dog, or a horse. Third, the trajectories cross regions with high estimated classification errors, suggesting a series of typical scenarios where an adversarial attack could be detected at these boundary crossings by a properly trained error/confidence estimator. More research is needed to study the exact structures herein visualized; perhaps the adversarial trajectories follow a pattern that could be detected and counter-attacked accordingly.

%

\subsection{A brief study on error estimate reliability}

\begin{figure*}[t]
	\centering
	\includegraphics[width=0.99\linewidth]{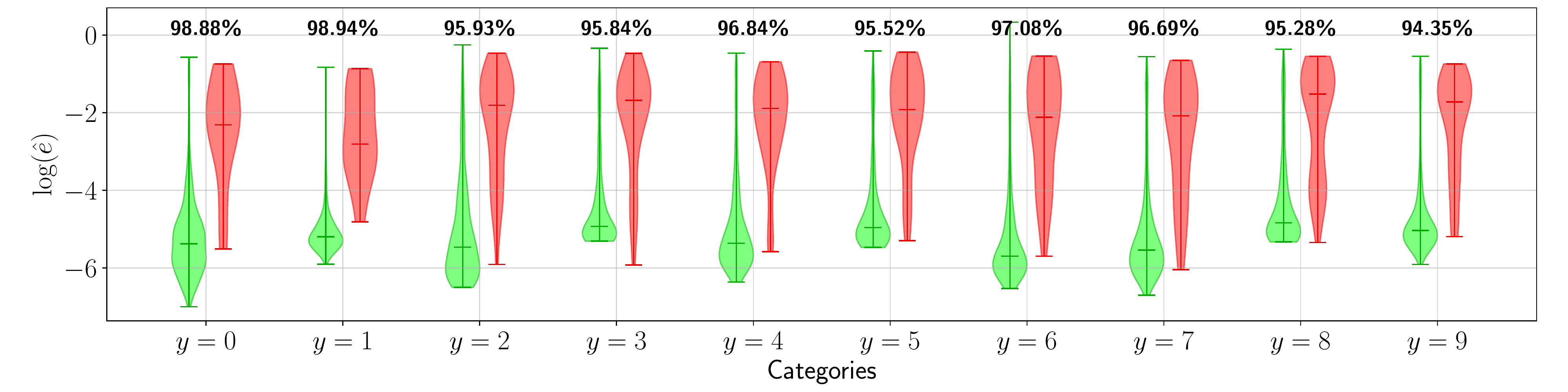}
	\includegraphics[width=0.99\linewidth]{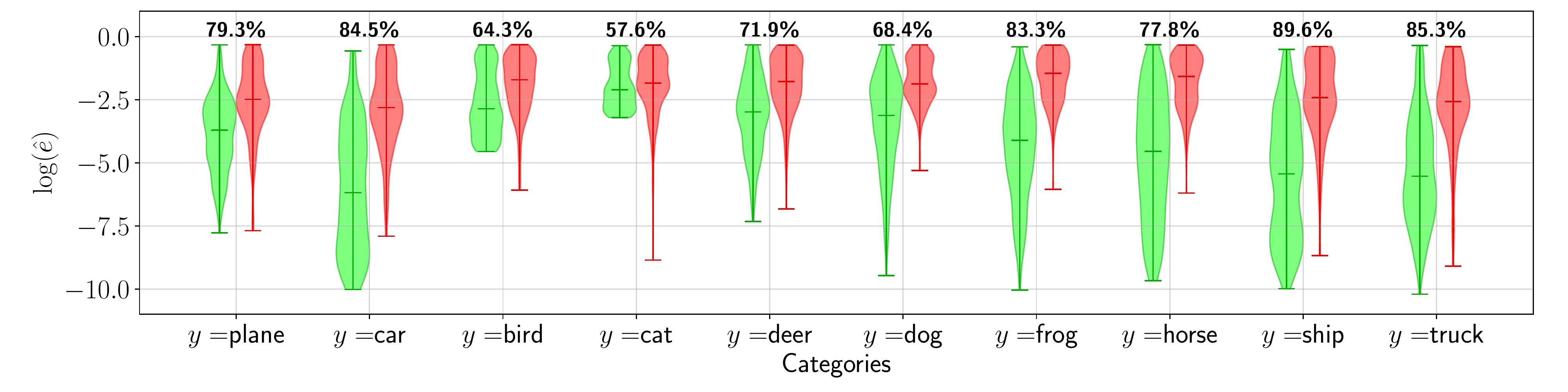}
	\caption{Violin plots showing the error estimate distributions on the test set of MNIST (top plot) and CIFAR-10 (bottom plot) for the self-introspective networks shown here. Each category is split into correctly classified samples (green violins) and misclassified samples (red violins). On top of each pair of violins for each output category, the sensitivity of the category is shown as well. Error estimates given by the self-introspective framework are higher for the majority of misclassified pixels in the test set in simpler datasets like MNIST, while in more complex data, such as CIFAR-10, less clear categories (with lower accuracies) remain with a higher expected error. With these results, $c=-\log_{10}(\hat{e})$ could be used as a metric of performance, and be employed in systems where reliability is required.}
	\label{fig:boxplots}
\end{figure*}

In some of our experiments, we have assumed that the output error estimates given by the self-introspective system should be reliable enough to detect most misclassified pixels. Such a claim needs to be verified quantitatively. To test this hypothesis, we have analyzed the output of the self-introspective module in the test set for both MNIST and CIFAR-10, on fully trained networks. Being in possession of the ground truth for the set, we have split the dataset into two categories: samples correctly classified by the network, and samples that have been misclassified. Then, the distributions of their respective error estimates are represented in green and red, respectively, by the violin plots of Figure \ref{fig:boxplots}. This result shows how misclassified images are, on average, two to three orders of magnitude worse than the correctly classified data for MNIST, in terms of expected classifier error. The case for CIFAR-10 is not as simple, and thus we can observe how the error is within $10^{-2}$ to $10^{-3}$ for key categories that are systematically misclassified, such as cats, birds and dogs. 

There are plenty of things to say about these plots. First, we can observe that a well-trained neural network can show a consistent difference between misclassified pixels and correct classifications in a dataset with consistent differentiability. Such is the case for the MNIST classifier, shown in the top plot, where the percent of correct classification for each of the individual categories is shown on top of each pair of violins. Unfortunately, there is more to this situation than meets the eye. In general, datapoints that are able to trespass the frontiers between two or more well-known regions of well-known neural network responses will appear to the estimator network as if they were legitimate classifications, responding with a low expected error. This phenomenon can be observed for CIFAR-10 in Figure \ref{fig:act_map}.(h), where many of the 'cat' and 'dog' categories have breached each other's limits, with plenty of datapoints infiltrating into the other response. In the violin plots, this can be seen in the shapes of the misclassified distributions, as a small fraction of the wrongly classified samples are evaluated as correct classification. It is also possible that the network we trained for CIFAR-10 is not optimally trained (our test set accuracy remains about 10\% below the state of the art \cite{Springenberg2014}, perhaps due to adding densely connected layers after the global averaging), and thus thanks to the self-introspective layer we can observe how the complete model could be improved for further training. This is in many ways a positive result: this network puts higher estimated errors in output categories where its performance is poorer, showing that it is aware that its expected performance will be worse than for other categories. Much more research is needed to better understand how these error estimation functions could be improved to detect abnormal neural network behavior, but for now we can observe, as a general rule, that consistently learned images can be observed as well-known responses in this new domain, whereas images that are less evident to the network will provide less typical classifier activations and, thus, a greater likelihood of producing an erroneous result. How this domain is handled, and how these error and confidence functions are defined, should be studied in the future.

\section{Conclusions}
\label{sec:conclusions}

In this experimental work, we have examined the idea of studying the behavior of hidden layers in a deep neural network, by means of unsupervised dimensionality reduction on records of hidden activation sequences. This was achieved by the use of a well-known network, a variational autoencoder, with which an interactive activation atlas can be produced. The autoencoder provides an additional layer of flexibility: it can be interrogated with new, incoming data, and additional networks and operations can be layered on top of its representations to further allow us to analyze and predict classifier behavior.

Understanding the underlying structure of a deep neural net is complicated, amongst other reasons, due to the difficulty in handling the vast number of connections, activations, and weights, and their final influence in the output layer. By applying nonlinear dimensionality reduction to the hidden layers, we can observe that a network responds very similarly to inputs that should be considered identical by the output layer. This, in turn, means that a network responds with an \textit{activation signature} when the input resembles a specific subset of the training data. Consequentially, misclassification is likely to occur whenever the sequence strays away from typical signatures. Well-known states for a network exist in a low-dimensional embedding, and are very sensitive to input perturbations that differ from those observed during training.

The potential of unsupervised dimensionality reduction as an essential part of model evaluation and design is not in itself a novel concept; see \cite{Andry2016,Olah2017,Olah2018,carter2019}. Studying a network with another network, on the other hand, does not seem to have been considered to date. For now, these experiments allows us to consider model evaluation as another interactive element residing within the model architecture. Self-introspection, understood as the addition of an autoencoder and secondary functions that can analyze and regulate the behavior of a given classifier, can also serve as a human interface for model reliability in more complex systems. For instance, in ensembles or assemblies of classifiers, the individual self-confidence of each network can be studied as the inverse of its expected error, $-\log_{10}(\hat{e}_i)$. Such a metric could be used to weigh the influence of each classifier to the output of the ensemble. Since the complete system is a neural network in itself, the activations for new incoming datapoints could be plotted in an activation map, and their expected accuracies could be estimated. Such an application could be helpful in the detection of adversarial attacks in complicated algorithms (e.g. face and voice recognition), where the autoencoder could serve as a private key: if the attack is subtle, but sufficient to trick the classifier, the activations may likely end up far away from well-known classification sequences, and thus be noticed by the autoencoder, whose gradient cannot be studied by the attacker, as long as its parameters are kept secret.

Whereas studying how the inputs elicit specific activations provides a bottom-up explanatory framework, applying unsupervised dimensionality reduction to the set of all possible activations inside a classifier results in somewhat of a top-down approach, where many interesting phenomena can be observed and analyzed as geometrical objects. This means, in turn, that current successful approaches in neural feature visualization could benefit from this approach. For example, results similar to those in Activation Atlases \cite{carter2019} could perhaps be redefined from a series of hierarchical, low-dimensional representations, by applying Feature Visualization at the bottleneck of an autoencoder that studies the activation space of the first convolutional layers of a network. Such a result could be achieved by modifying the input so that the hidden activation pattern is located at a specific point in activation space. Other methods, such as Neural Style Transfer \cite{Gatys2015ANA}, could study the location of specific artistic styles as locations in the activation subspace of the first convolutional layers of a pretrained model, and tune the final result in this low-dimensional representation to their particular interests. Activations in recurrent neural networks (RNNs) could be encoded in a low-dimensional space, showing typical and atypical trajectories, and possibly enforcing corrections not on their input/output data, but instead on the output of the autoencoder, by rewarding specific trajectories and discarding others. 

In general, applying interactive, explicit dimensionality reduction methods opens up a window of explainability that is easy to replicate (only an unsupervised neural network is needed) and whose output is easily exploitable in further calculations and training processes. More research is required in order to understand the reliability, structure and significance of these representations and how they can aid us in more complex problems.

\section*{Acknowledgments}

The authors would like to thank Samuel S. Streeter and Benjamin W. Maloney from Thayer School of Engineering at Dartmouth College (03755 Hanover, NH) for the multiple fruitful discussions regarding this work and its potential future applications.

Research reported in this article was funded by projects R01 CA192803 and F31 CA196308 (National Cancer Institute, US National Institutes of Health), FIS2010-19860 (Spanish Ministry of Science and Innovation), TEC2016-76021-C2-2-R (Spanish Ministry of Science, Innovation and Universities), DTS17-00055 (Spanish Minstry of Economy, Industry and Competitiveness and Instituto de Salud Carlos III), INNVAL 16/02 (IDIVAL), and INNVAL 18/23 (IDIVAL), as well as PhD grant FPU16/05705 (Spanish Ministry of Education, Culture, and Sports) and FEDER funds.

\bibliographystyle{unsrt}  
\bibliography{hRepr}

\end{document}